\documentclass[conference]{IEEEtran}
\IEEEoverridecommandlockouts
\usepackage{cite}
\usepackage{amsmath,amssymb,amsfonts}
\usepackage{graphicx}
\usepackage{textcomp}
\usepackage{xcolor}
\usepackage{algorithm}         
\usepackage{algpseudocode}      
\usepackage{amsmath}           
\usepackage{amssymb}            
\usepackage{amsfonts}         
\usepackage{float}   
\usepackage{booktabs} 
\usepackage{amssymb}  
\usepackage{tabularx} 
\usepackage{booktabs}
\usepackage{multirow}
\def\BibTeX{{\rm B\kern-.05em{\sc i\kern-.025em b}\kern-.08em
    T\kern-.1667em\lower.7ex\hbox{E}\kern-.125emX}}
\begin{document}

\title{Multi-strategy Improved Northern Goshawk Optimization for WSN Coverage Enhancement\\
{\footnotesize \textsuperscript{}}
\thanks{Identify applicable funding agency here. If none, delete this.}
}

\author{\IEEEauthorblockN{1\textsuperscript{st} Yiran Tian}
\IEEEauthorblockA{\textit{Mudanjiang Normal University}\\
China, Heilongjiang Province \\{157000}\\
tyr199810@163.com}
\and
\IEEEauthorblockN{2\textsuperscript{nd}Yuanjia Liu}
\IEEEauthorblockA{\textit{Xi'an University of Technology}\\
China, Shaanxi Province \\{710000}\\
minato\_liu@outlook.com}}

\maketitle

\begin{abstract}
To enhance the coverage rate of Wireless Sensor Networks (WSNs), this paper proposes an advanced optimization strategy based on a multi-strategy integrated Northern Goshawk Optimization (NGO) algorithm. Specifically, multivariate chaotic mapping is first employed to improve the randomness and uniformity of the initial population. To further bolster population diversity and prevent the algorithm from stagnating in local optima, a bidirectional population evolutionary dynamics strategy is incorporated following the pursuit-and-evasion phase, thereby facilitating the attainment of the global optimal solution. Extensive simulations were conducted to evaluate the performance of the proposed multi-strategy NGO in WSN coverage. Experimental results demonstrate that the proposed algorithm significantly outperforms existing benchmarks in terms of both coverage enhancement and node connectivity.
\end{abstract}


\section{Introduction}
Wireless Sensor Networks (WSNs) are distributed intelligent systems comprised of a vast number of miniature, low-power nodes equipped with sensing, computing, and wireless communication capabilities. These nodes typically form a network through self-organization and multi-hop data transmission\cite{b1}\cite{b2}. Each sensor node collaboratively perceives, collects, processes, and transmits data regarding target objects within a monitored region to a central sink node for user analysis and decision-making.

The performance of such networks is fundamentally determined by their monitoring coverage \cite{b3}. However, practical deployments often suffer from uneven node distribution, which leads to sensing redundancies and coverage holes, ultimately resulting in a significant waste of hardware resources\cite{b4}\cite{b5}. Consequently, optimizing coverage has emerged as one of the most critical challenges in the field of Wireless Sensor Network (WSN) research. 

In WSNs, achieving optimal network coverage is categorized as an NP-hard problem, implying that finding an exact solution is computationally intensive and often intractable within a reasonable timeframe \cite{b6}. Given this complexity, Swarm Intelligence (SI) optimization algorithms have emerged as a robust alternative for identifying high-quality approximate solutions. Inspired by natural processes, SI algorithms have proven to be promising tools for addressing multifaceted challenges in WSNs. These metaheuristics demonstrate particular efficacy in event detection and query processing, thereby enhancing the precision and reliability of identifying real-world events within the monitoring domain \cite{b7}\cite{b8}.

The suitability of SI algorithms for NP-hard problems stems from their ability to efficiently explore vast and complex search spaces. Ensuring the accuracy and reliability of data collected by sensor nodes involves various intricate factors, including protocol design, the integration of anomaly detection techniques, and the application of bio-inspired heuristics to streamline data processing \cite{b9}. Through these synergistic optimization efforts, SI algorithms facilitate the generation of credible data, ultimately augmenting the overall performance and practical utility of WSNs.

To address the challenges of coverage optimization in WSNs, various enhanced swarm intelligence metaheuristics have been developed and deployed. For instance, Toloueiashtian et al. \cite{b10} introduced an improved Whale Optimization Algorithm (WOA) tailored for the point coverage problem. By refining the three core mechanisms of WOA—exploration, spiral attack, and bubble-net searching—this approach significantly boosts the coverage rate by identifying optimal node configurations. Similarly, Akram et al. \cite{b11} proposed a strategy utilizing adaptive learning automata. By equipping sensor nodes with autonomous learning capabilities, the network can dynamically select appropriate states at any given moment, thereby optimizing overall coverage performance.

Addressing the specific issues of high deployment costs and insufficient effective coverage, Ou et al.\cite{b12} developed a multi-strategy Grey Wolf Optimizer (GWO). This method incorporates several refinement strategies to help the algorithm circumvent premature convergence and escape local optima. Furthermore, to mitigate coverage blind spots and redundancy during the random deployment of Heterogeneous Wireless Sensor Networks (HWSNs), Cao et al. \cite{b13} presented an optimization strategy based on an enhanced Social Spider Optimization (SSO) algorithm. Specifically, they integrated chaotic initialization to accelerate global convergence and refined neighborhood search, global search, and matching radius mechanisms to bolster search efficiency. This approach not only enhances network coverage but also effectively reduces energy consumption.

In addition to meeting the sensing requirements of the monitored region, the design of coverage-oriented WSNs must rigorously account for energy consumption. Consequently, a significant body of research seeks to prolong network operational life while maintaining high coverage quality \cite{b14}. For instance, Yarinezhad and Hashemi \cite{b15} developed a sensor deployment methodology for the target coverage problem, leveraging two enhanced variants of Particle Swarm Optimization (PSO) to simultaneously maximize coverage and network longevity.

To address the limitations of conventional coverage models, Elhoseny et al. \cite{b16} employed Genetic Algorithms (GA) to optimize WSN coverage. Their approach ensures continuous monitoring of specified targets for the maximum possible duration under constrained energy resources, resulting in marked improvements in both network lifetime and throughput. Furthermore, some researchers have integrated area-coverage mitigation techniques with efficient clustering methodologies to optimize energy utilization. A notable example is the Area Coverage-Aware Clustering Protocol (ACACP), which optimizes power consumption across sensor activation, network clustering, and multi-hop communication phases to extend the overall system lifespan without compromising sensing range \cite{b17}.

The Northern Goshawk Optimization (NGO) algorithm, inspired by the predatory behavior of goshawks, has demonstrated significant potential in tackling complex optimization problems due to its robust search capabilities and parallel processing potential \cite{b18}. Despite these advantages, the standard NGO still encounters several critical bottlenecks, including a susceptibility to local optima, a lack of initial population diversity stemming from stochastic initialization, and an inherent imbalance between exploration and exploitation.

To mitigate these limitations, several enhanced variants of the NGO algorithm have been proposed. For instance, Liang et al.\cite{b19} introduced the Enhanced NGO (ENGO) by integrating polynomial interpolation strategies with diverse opposition-based learning methods. This hybrid approach effectively maintains a dynamic equilibrium between exploration and exploitation, facilitating faster convergence toward high-quality solutions for high-dimensional problems. Similarly, Sadeeq and Abdulazeez \cite{b20}  proposed an innovative mode-switching mechanism between exploration and exploitation phases, augmented by Levy flight to bolster global search capabilities and prevent stagnation in local optima. Furthermore, Zeng et al. \cite{b21}  developed a multi-strategy improved NGO tailored for global optimization and engineering design, specifically targeting the issues of premature convergence and localized trapping.

To overcome the inherent susceptibility of the NGO algorithm to local extrema and further enhance the coverage performance of WSN models, this paper introduces a multi-strategy Improved Northern Goshawk Optimization (INGO) algorithm. The proposed INGO framework incorporates multivariate chaotic mapping for population initialization and integrates a bidirectional population evolutionary dynamics strategy. These enhancements are specifically designed to bolster global search efficiency and ensure a robust balance between exploration and exploitation.

By leveraging the INGO algorithm, we optimize both the coverage rate and node connectivity of WSN models. To rigorously validate the superiority of the proposed approach, extensive comparative experiments were conducted. The performance of INGO was benchmarked against the standard NGO, the Artificial Bee Colony (ABC) algorithm \cite{b22}, an Improved Wild Horse Optimizer (IWHO) \cite{b23}, and a Firefly Algorithm (FA)-based deployment method \cite{b24}. Experimental results demonstrate that the INGO algorithm effectively yields superior coverage optimization and connectivity, consistently outperforming the aforementioned state-of-the-art metaheuristics.

The remainder of this paper is organized as follows: Section II describes the WSN coverage model and the problem formulation; Section III details the proposed INGO algorithm and its constituent strategies; Section IV presents the experimental results and comparative analysis; finally, Section V concludes the paper and discusses future research directions.
\section{Related work}
\subsection{System Modeling and Problem Formulation}
Assume that the monitoring region is a two-dimensional (2D) plane of size $L\times M$, where $N$ homogeneous sensor nodes are stochastically deployed. In this WSN architecture, the sensing range of each node is modeled as a circular region centered at the node's position. Let $R$ denote the sensing radius and $R_c$ represent the communication radius, with the operational constraint $R_c\geq 2R$ ensuring seamless network connectivity. The set of sensor nodes is defined as $S={{}\left\{ s_1,s_2,s_3....s_n \right\}}$.

Suppose a sensor node $v_i$ is located at $(x_i,y_i)$ and a target point $u_j$ is situated at $(x_j,y_j)$. The Euclidean distance between the node $v_i$ and the target $u_j$, denoted as $d(v_i,u_j)$, is calculated as follows \cite{b25}:

\begin{equation}
d(v_i,u_j)=\sqrt{(x_i-x_j)^2+(y_i-y_j)^2}\label{eq}
\end{equation}
In accordance with the Boolean sensing model, if target $u_j$ falls within the circular sensing range of node $v_i$, the sensing quality is $1$; otherwise, it is $0$. Consequently, the sensing probability $p(v_i,u_j)$ of node $v_i$ relative to target $u_j$ is defined as:

\begin{equation}
p(v_i,u_j)=\left\{\begin{matrix} 
  1, d(v_i,u_j)\le  R\\  
  0, d(v_i,u_j) > R\
\end{matrix}\right.\label{eq}
\end{equation}

In practical scenarios, a single target within the monitoring area may be perceived by multiple sensors simultaneously. To account for this, the individual sensing probabilities are aggregated into a joint sensing probability, defined as:

\begin{equation}
p(v_i,u_j)=1-\prod_{i=1}^{N}[1-p(v_i,u_j)]\label{eq}
\end{equation}

where $N$ denotes the total number of sensor nodes deployed within the region.

Coverage rate serves as a fundamental metric for evaluating the performance of WSNs \cite{b26}. Structurally, the coverage rate is defined as the ratio of the effectively covered area to the total monitoring area. By discretizing the region into a grid of points, the overall coverage rate Cov can be formulated as follows:

\begin{equation}
Cov=\frac{\sum_{j=1}^{L\times M}{p(v_i,u_j)}}{L\times M}\label{eq}
\end{equation}

In this study, $Cov$ in Equation (4) is utilized as the objective function, which the proposed improved algorithm aims to maximize to identify the optimal node configuration.

\begin{figure}[ht]
  \centering
  \includegraphics[width=0.8\linewidth]{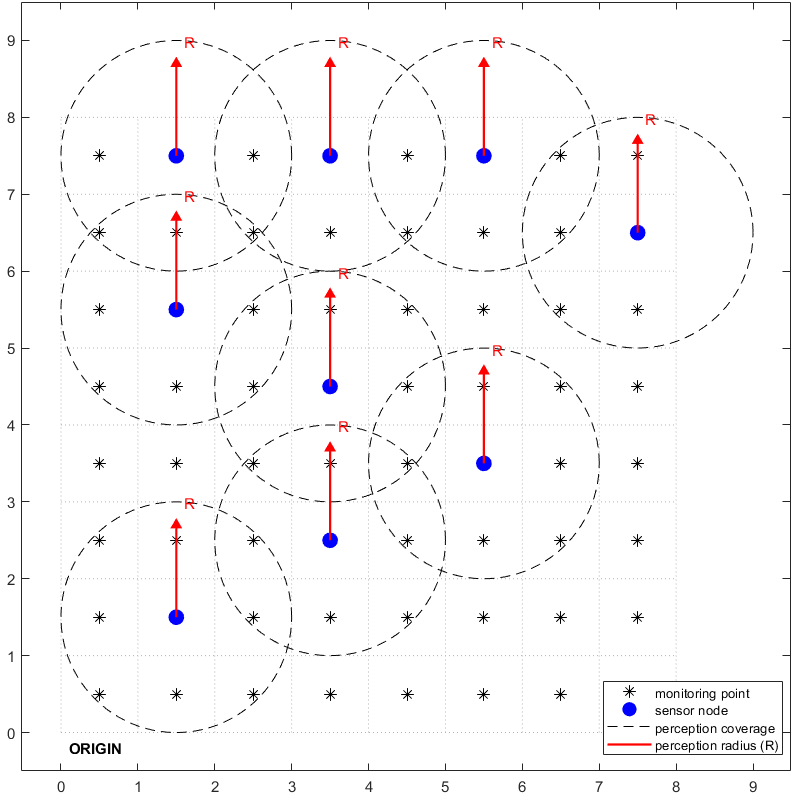}
  \caption{Wireless Sensor Network Coverage}
  \label{fig:wsn_coverage}
\end{figure}

To evaluate the coverage performance of the WSN, a numerical simulation environment is established as illustrated in Fig.~\ref{fig:wsn_coverage}. The simulation process is detailed as follows:

Area Discretization: The monitored region is partitioned into a uniform grid of equal-sized cells, with a discrete monitoring point (represented by a black star in the simulation) situated at the center of each grid cell.

Coverage Criterion: A monitoring point is considered "covered" if and only if its Euclidean distance to the nearest sensor node is less than or equal to the sensing radius $R$.

Approximation Accuracy: The overall network coverage rate is computed as the ratio of the cumulative area of all covered cells to the total area of the monitoring region.

Convergence: As the grid granularity increases (i.e., finer discretization), the calculated coverage rate asymptotically approaches the true physical coverage level of the network.

\subsection{Standard Northern Goshawk Optimization}
The Northern Goshawk Optimization (NGO) algorithm \cite{b27} is a population-based metaheuristic inspired by the strategic hunting behaviors of goshawks. The search agents, representing individual goshawks, navigate the solution space through a two-phase process:

Phase 1: Prey Identification and Rapid Strike: This stage simulates the initial detection and swift attack on prey, primarily facilitating global exploration to identify promising regions within the search space.

Phase 2: Chase and Escape: This stage models the subsequent pursuit when prey attempts to flee, focusing on fine-grained local exploitation to drive the population toward the global optimum.

\subsubsection{Phase 1: Prey Identification and Strike (Exploration)}
During the first phase, a northern goshawk randomly selects a prey and initiates a rapid strike. By stochastically choosing prey across the entire search space, the NGO algorithm executes an extensive global search to locate the optimal region. This stochastic behavior is mathematically formulated as follows:

\begin{equation}
P_i=X_k,i=1,2,...,N, \\k=1,2,...i-1,i+1,...N\label{eq}
\end{equation}

\begin{equation}
x_{i,j}^{new,P_1}=\left\{\begin{matrix} 
  x_{i,j}+r(p_{i,j}-I_{x_{i,j}}), F_{P_i}<  F_i\\  
  x_{i,j}+r(x_{i,j}-p_{x_{i,j}}), F_{P_i}\geq F_i\
\end{matrix}\right.\label{eq}
\end{equation}

\begin{equation}
X_i=\left\{\begin{matrix} 
  X_i^{new,P_1}, F_i^{new,P_1}<  F_i\\  
  X_i, F_i^{new,P_1} \geq F_i\
\end{matrix}\right.\label{eq}
\end{equation}
In Equation (5), $P_i$ denotes the location of the prey targeted by the $i-th$ goshawk, while $X_k$ represents the position of the $k-th$ goshawk within the population. Here, $N$ signifies the total population size, and k is a randomly selected integer index from the interval $[1,N]$.

Regarding Equation (6), $x_{i,j}$ and $x_{i,j}^{new,P1}$ denote the current and updated positions of the $i-th$ goshawk in the $j-th$ dimension, respectively. The terms $F_{P_i}$ and $F_i$ represent the objective function values (fitness) associated with the prey's position and the $i-th$ goshawk's current position. The parameter $r$ is a random scalar uniformly distributed in the range $[0,1]$, and $I$ is a stochastic integer parameter assigned a value of either $1$ or $2$.

Finally, in Equation (7), $X_i^{new,P_1}$ represents the proposed new position for the $i-th$ goshawk, and $F_i^{new,P_1}$ corresponds to its updated fitness value. The algorithm adopts a greedy selection mechanism, where the new position is accepted only if it yields an improvement in the objective function value.

\subsubsection{Phase 2: Chase and Escape (Exploitation)}

\begin{figure}[ht]
  \centering
  \includegraphics[width=0.8\linewidth]{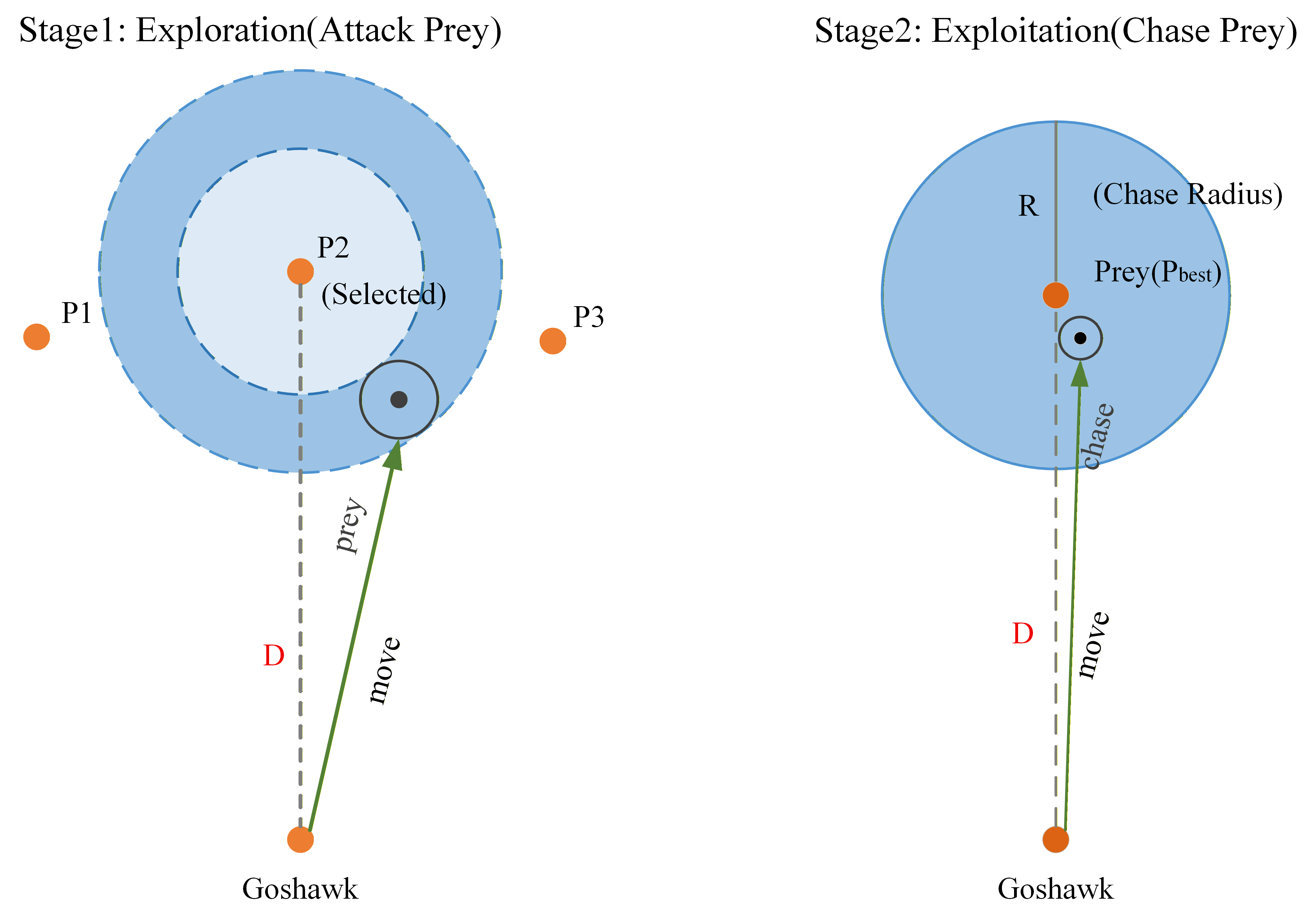}
  \caption{NGO algorithm exploration and pursuit phase diagram}
  \label{fig2}
\end{figure}

In the second phase, once the prey attempts to flee upon being targeted, the northern goshawk initiates a high-speed pursuit, culminating in the capture of the prey. This exploitation behavior is characterized by a localized search around the current position, as illustrated in the schematic diagram of Fig.~\ref{fig2}. The mathematical model for this phase is defined by Equations (8) through (10):

\begin{equation}
x_{i,j}^{new,P_2}=x_{i,j}+R(2\times r-1)x_{i,j}\label{eq}
\end{equation}

\begin{equation}
R=0.02\times(1-\frac{t}{T})\label{eq}
\end{equation}

\begin{equation}
X_i=\left\{\begin{matrix} 
  X_i^{new,P_2}, F_i^{new,P_2}<  F_i\\  
  X_i, F_i^{new,P_2} \geq F_i\
\end{matrix}\right.\label{eq}
\end{equation}

In these equations, $x_{i,j}^{new,P_2}$ represents the newly proposed coordinate of the $i-th$ goshawk in the $j-th$ dimension during the chase-and-escape phase. Accordingly, $X_i^{new,P_2}$ denotes the updated position vector for the $i-th$ goshawk, and $F_i^{new,P_2}$ corresponds to its resultant objective function value.

The parameter $R$ signifies the radius of the exploitation neighborhood, which dynamically decreases as the simulation progresses. Specifically, $t$ denotes the current iteration count, while $T$ represents the maximum number of iterations allowed. This time-varying radius $R$ facilitates a transition from broader local searching to refined convergence as the algorithm approaches the global optimum.

\section{The Proposed Improved NGO (INGO) Algorithm}

While the standard NGO algorithm exhibits superior convergence performance compared to several other swarm intelligence metaheuristics, it remains susceptible to being trapped in local extrema\cite{b28}. To address this limitation and further enhance the coverage performance of WSNs, we propose the Improved Northern Goshawk Optimization (INGO) algorithm. The INGO framework introduces two synergistic strategies:

\begin{itemize}
\item Diverse Chaotic Map Initialization : This strategy is employed to initialize the population, thereby significantly increasing population diversity and expanding the initial search scope.

\item Bidirectional Population Evolutionary Dynamics: This novel strategy is incorporated to facilitate the algorithm's escape from local optima, enabling a more profound exploration of the global optimal solution.

The primary objective of the INGO algorithm is to effectively bolster the coverage efficiency and network connectivity of WSNs, offering a robust methodology for large-scale sensor deployment challenges.
\end{itemize}

\subsection{Initialization via Diverse Chaotic Map Initialization Strategy (DCMIS)}

In the standard NGO algorithm, the initial population is typically generated using a one-time pseudo-random function. Empirical evidence suggests that this stochastic approach often leads to "clustering" of individuals within the solution space, causing the algorithm to originate from suboptimal starting points and subsequently diminishing both global exploration and local exploitation capabilities\cite{b29}. To mitigate this, the INGO algorithm adopts the Diverse Chaotic Map Initialization Strategy (DCMIS). By alternating and coupling Logistic\cite{b30} and Sine\cite{b31} maps, this strategy generates chaotic sequences characterized by low correlation and high ergodicity. This ensures that the initial individuals are uniformly distributed across the entire feasible domain, thereby enhancing the diversity and coverage of the search starting points.

The mathematical formulation of the DCMIS is presented in Equations (11) through (13):

\begin{equation}
z_{seed}=rand(N,dim)\label{eq}
\end{equation}

\begin{equation}
z=sin(\pi\times(z_{seed})\times(1-z_{seed})+sin(\pi\times z_{seed})))\label{eq}
\end{equation}

\begin{equation}
X=Lowerbound+\left| z \right|\times(Upperbound-Lowerbound)\label{eq}
\end{equation}

In Equation (11), $z_{seed}$ denotes the random seed matrix of size $N\times dim$, serving as the foundational input for the chaotic sequence, where $N$ signifies the population size.

Equation (12) represents the integration of the Logistic and Sine maps. This formulation builds upon the Logistic map by introducing additional complexity and non-linearity, where z represents the output chaotic value matrix after applying these sophisticated non-linear transformations to $z_{seed}$.

Finally, in Equation (13), $X$ represents the matrix containing the final initial positions of the population. While the values generated by the multivariate chaotic mapping appear stochastic, they effectively cover the search space with superior uniformity. This produces a high-quality, diverse initial population, providing a more robust foundation for the algorithm's subsequent optimization phases.

\subsection{Bidirectional Population Evolutionary Dynamics (BPED)}

To assist the NGO algorithm in escaping local optima during the mid-to-late stages of iteration and to bolster population vitality, a Bidirectional Population Evolutionary Dynamics (BPED) strategy is integrated following the chase-and-escape phase.

The BPED strategy is primarily inspired by the Pareto Principle (the 80/20 rule) observed in natural systems. Specifically, it retains the top 20\% of individuals with the highest fitness values and subjects them to controlled natural variation to maintain elite traits while exploring adjacent regions. The mathematical mechanism for this process is detailed in Equations (14) through (17):

\begin{equation}
w=\frac{1}{2}\times  (\sin (2\pi\times freq\times t +\pi )\pi\times  (\frac{t}{T_{max}} )+1)\label{eq}
\end{equation}

\begin{equation}
freq=\frac{1}{dim}\label{eq}
\end{equation}

\begin{equation}
z=\sin (\pi\times(z_{rand}\times (1-z_{rand})+\sin (\pi \times z_{rand})))\label{eq}
\end{equation}

\begin{equation}
X_{good,new}=X_q+w\times (x_{best}-round(1+z)\times X_k)\label{eq}
\end{equation}

In these formulations, $w$ represents a dynamic weight factor designed to modulate the search behavior. During the initial stages of the algorithm, $w$ provides a stable weight to facilitate steady convergence toward the target region. In the latter stages, $w$ introduces stochastic oscillations to enhance population diversity, thereby empowering the algorithm to circumvent local stagnation.

Furthermore, $t$ denotes the current iteration count, $T_{max}$ signifies the maximum number of iterations, and dim represents the dimensionality of the optimization problem. The term freq defines the oscillation frequency, while $z$ represents the chaotic disturbance derived from the nonlinear transformation of a random seed $z_{rand}$. The resulting vector $X_{good,new}$ denotes the updated position of the elite individuals, guided by the global best position $x_{best}$ and a randomly selected individual $X_k$ to preserve evolutionary momentum.

In Equation (16), $z$ represents the chaotic disturbance value. By leveraging the ergodicity of chaos, this parameter generates more effective perturbations that facilitate a comprehensive exploration of the search space. Unlike the standard rand function, which may exhibit clustering or repetitive values during finite iterations despite its long-term uniform distribution, $z_{rand}$ is employed here as a specialized stochastic seed to ensure superior independence and non-periodicity in the generated sequences.

Equation (17) defines the updated position of an "elite individual," denoted as $X_{good,new}$, which signifies the target coordinate calculated for the subsequent iteration. The mechanism is governed by the following components:

Baseline Vector ($X_q$): This represents the position of a randomly selected elite individual q, serving as the fundamental reference for the current update.

Global Guidance ($x_{best}$): This denotes the global optimal position—the individual with the highest fitness found by the entire population thus far—which exerts an attractive force on $X_q$ toward the known global target.

Chaotic Strategy Switch ($round(1+z)$): Controlled by the chaotic value $z$, this term functions as a stochastic operator that yields discrete values of either $1$ or $2$. It introduces non-linear scaling to the update process.

Repulsive Vector ($X_k$): By subtracting the position (or twice the position) of another randomly selected elite individual $k$ from $x_{best}$, the strategy introduces a repulsive component in the update direction. This prevents the population from blindly converging toward a single point ($x_{best}$), thereby enhancing exploratory capacity and effectively circumventing premature convergence.

he second component of the BPED strategy targets the remaining 20\% of the population characterized as "non-elite" individuals. These agents are assumed to be stagnant within unproductive regions of the solution space. To prevent the algorithm from wasting computational resources on these suboptimal solutions, a forced re-exploration mechanism is implemented to uncover new possibilities. Specifically, a stochastic dual-path update is applied to these individuals based on a random threshold:

\begin{itemize}
\item Case 1: Local Refinement via Dynamic Boundaries (rand<0.5) In the first scenario, the stagnant individual is transformed into a new exploratory agent that performs a fine-grained search in the vicinity of the current global optimum, $x_{best}$. This process is mathematically formulated as follows:

\begin{equation}
\begin{split}
X_{bad,new} = {} & x_{best} + \text{sign}(\text{rand}-0.5) \\
& \times (lb_{ap} + \text{rand} \times (ub_{ap} - lb_{ap}))
\end{split}
\label{eq:bad_new_position}
\end{equation}

In Equation (18), $X_{bad,new}$ denotes the updated position of the non-elite individual. The term $sign(rand-0.5)$ serves as a directional perturbation vector. Notably, $ub_{ap}$ and $lb_{ap}$ represent dynamic boundaries that progressively narrow as the iteration count $t$ increases, facilitating a transition toward more focused exploitation.

\item Case 2: Large-Scale Stochastic Mutation ($rand\geq0.5$) In the remaining 50\% of cases, the algorithm executes a radical stochastic mutation to propel the individual toward entirely unexplored regions of the search space. This mechanism is critical for circumventing persistent local optima:

\begin{equation}
\begin{split}
X_{bad,new} = {} & X_{z1} - 2 \times \text{sign}(\text{rand}-0.5) \\
& \times (lb + \text{rand} \times (ub - lb))
\end{split}
\label{eq:bad_new_mutation}
\end{equation}

In Equation (19), $X_{z1}$ represents the current coordinates of the targeted non-elite individual, while ub and lb signify the global upper and lower bounds of the search space, respectively. The multiplier $2\times sign(rand-0.5)$ acts as an amplified perturbation vector, ensuring that the mutation step is sufficiently large to displace the agent from its current suboptimal basin.
\end{itemize}

\subsection{Implementation of the INGO Algorithm for WSN Coverage}

\begin{figure}[ht]
  \centering
  \includegraphics[width=\linewidth]{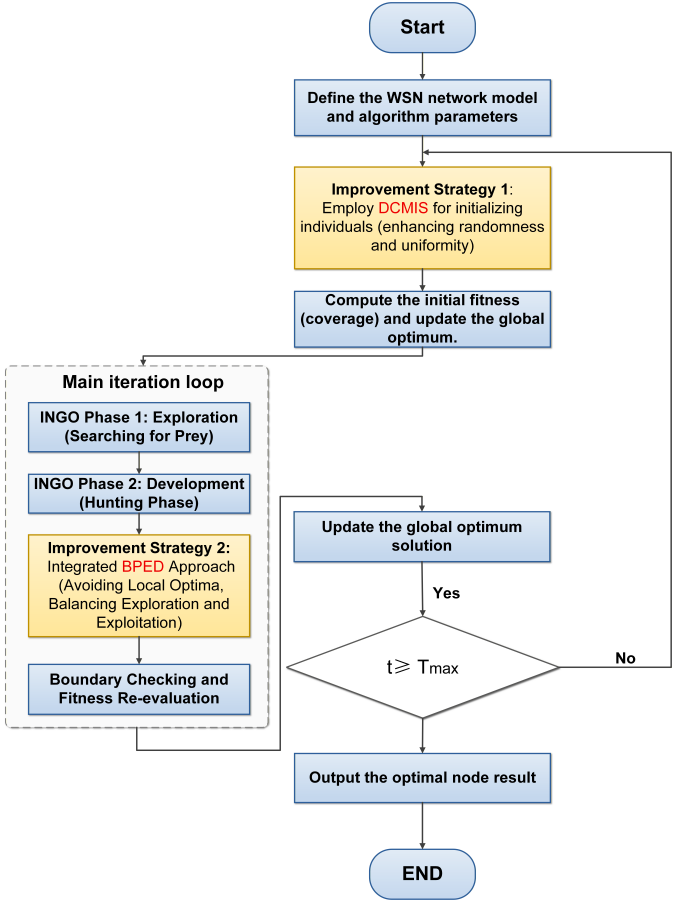}
  \caption{INGO-WSN Coverage Overall Flowchart}
  \label{fig3}
\end{figure}

The integration of the proposed INGO algorithm into the WSN coverage optimization model follows a structured in Fig.~\ref{fig3} as below:

\begin{algorithm}[H]
\caption{Improved Northern Goshawk Optimization (INGO)}
\label{alg:ingo}
\begin{algorithmic}[1]
\Require $f_{obj}$ (Objective function), $lb, ub, N, T$
\Ensure $x_{best}$ (Global optimum), $Cov$ (Optimal coverage rate)
\State \textbf{Initialization:} Generate population $X$ via DCMIS strategy using Eq. (11)--(13)
\State Evaluate initial fitness and identify the global best $x_{best}$
\For{$t = 1$ \textbf{to} $T$}
    \State \textbf{// Stage I: Standard NGO Metaheuristic}
    \For{$i = 1$ \textbf{to} $N$}
        \State \textbf{Exploration:} Identify prey and initiate strike  Eq. (5)--(6)
        \State \textbf{Exploitation:} Execute chase-and-escape pursuit  Eq. (8)--(9)
        \State \textbf{Update:} Apply greedy selection for position $X_i$  Eq. (7) \& (10)
    \EndFor
    \State \textbf{// Stage II: Bidirectional Population Evolution (BPED)}
    \State Rank $X$ and partition into $X_{elite}$ (Top 20\%) and $X_{non\text{-}elite}$ (Bottom 20\%)
    \State \textbf{Evolution of $X_{elite}$:} Update $X_{elite}$ Eq. (17) using $w$ and $z$ from Eq. (14)--(16)
    \State \textbf{Re-exploration of $X_{non\text{-}elite}$:}
    \For{each $x \in X_{non\text{-}elite}$}
        \If{$\text{rand} < 0.5$}
            \State Local search with dynamic boundaries Eq. (18)
        \Else
            \State Large-scale stochastic mutation Eq. (19)
        \EndIf
    \EndFor
    
    \State Update $x_{best}$, $f_{best}$, and $Convergence\_curve(t)$
\EndFor

\State \Return $x_{best}, Convergence\_curve$
\end{algorithmic}
\end{algorithm}

Step 1: Model Construction and Parameter Initialization. The WSN coverage optimization framework is established by defining essential network parameters, including the dimensions of the monitoring area ($L\times M$), the total number of sensor nodes ($N$), sensing radius ($R$), communication radius ($R_c$), and the discretization granularity of the grid points.

Step 2: Population Initialization via DCMIS. The initial population of goshawks is generated using the DCMIS. This step ensures that the starting agents possess high diversity and are uniformly distributed across the search space, providing a robust foundation for global exploration.

Step 3: Standard NGO Iteration Phase. The primary iteration loop is executed, where search agents update their positions through the simulated predatory behaviors of northern goshawks. This phase encompasses both global exploration (Prey Identification and Strike) and local exploitation (Chase and Escape).

Step 4: Enhanced Optimization via BPED Strategy. Following the standard NGO update, the BPED is invoked. The population is ranked by fitness, where the top 20\% elite individuals evolve toward the global best position to refine accuracy, while the bottom 20\% stagnant individuals undergo stochastic mutation or re-initialization to circumvent local optima.

Step 5: Convergence and Performance Evaluation. Upon satisfying the termination criteria, the optimal solution vector is extracted to determine the final coordinates of the sensor nodes. The network coverage rate is then computed via the objective function. Finally, convergence curves and node deployment layouts are generated to verify the efficacy of the optimization.

The comprehensive flowchart for the INGO-based WSN coverage optimization is illustrated in Figure 3, and the detailed procedural execution is summarized in Algorithm 1.

\section{Experimental Results and Discussion}

To validate the efficacy of the proposed INGO algorithm and its optimization performance within the WSN coverage model, two categories of experiments were conducted: Ablation Studies to verify the contribution of individual strategy components, and Network Performance Experiments to evaluate the algorithm's application in practical WSN scenarios. All simulations were performed on a Windows 10 platform utilizing the MATLAB R2024a environment.

\subsection{Comparative Testing of the Improved Algorithm}

To systematically analyze the effectiveness of the proposed DCMIS and BPED strategies, an ablation study was designed as detailed in Table 1. This comparison evaluates the incremental performance gains provided by each enhancement.

For a comprehensive assessment of the algorithm's convergence and robustness, $15$ benchmark test functions were selected (as listed in Table 2). To mitigate the impact of stochastic variation, each algorithm was executed independently for 20 trials with a maximum iteration limit of $500$. The performance metrics, including the Best Score, Mean Score, and Standard Deviation (Std Dev), are recorded and summarized in Table 3.

\begin{table}[htbp]
\centering
\caption{Design of the ablation study for strategy verification.}
\label{tab:ablation_design}
\begin{tabularx}{\columnwidth}{@{} lcc X @{}}
\toprule
\textbf{Algorithm} & \textbf{DCMIS} & \textbf{BPED} & \textbf{Verification Purpose} \\ 
\midrule
NGO        & $\times$      & $\times$      & Baseline for overall assessment. \\ \addlinespace
INGO-DCMIS & \checkmark    & $\times$      & Evaluates initial solution quality and global ergodicity. \\ \addlinespace
INGO-BPED  & $\times$      & \checkmark    & Evaluates local optima avoidance and accuracy. \\ \addlinespace
INGO       & \checkmark    & \checkmark    & Validates the synergistic effect of both strategies. \\ 
\bottomrule
\end{tabularx}
\end{table}

\begin{table}[htbp]
\centering
\caption{Mathematical definitions of the 15 benchmark test functions.}
\label{tab:benchmark_functions}
\footnotesize 
\begin{tabularx}{\columnwidth}{@{} l X c c @{}}
\toprule
\textbf{Name} & \textbf{Function Formulation} & \textbf{Dim} & \textbf{Range} \\ 
\midrule
$F_1$ & $\sum_{i=1}^{n}x_i^2$ & 30 & $[-30, 30]$ \\
$F_2$ & $\sum_{i=1}^{n}|x_i| + \prod_{i=1}^{n}|x_i|$ & 30 & $[-10, 10]$ \\
$F_3$ & $\sum_{i=1}^{n}(\sum_{j=1}^{i}x_j)^2$ & 30 & $[-100, 100]$ \\
$F_4$ & $\max_{i}\{|x_i|, 1 \le i \le n\}$ & 30 & $[-100, 100]$ \\
$F_5$ & $\sum_{i=1}^{n-1}[100(x_{i+1}-x_i^2)^2 + (x_i-1)^2]$ & 30 & $[-30, 30]$ \\
$F_6$ & $\sum_{i=1}^{n}(\lfloor x_i + 0.5 \rfloor)^2$ & 30 & $[-100, 100]$ \\
$F_7$ & $\sum_{i=1}^{n}ix_i^4 + \text{rand}(0,1)$ & 30 & $[-1.28, 1.28]$ \\
$F_8$ & $\sum_{i=1}^{n}-x_i\sin(\sqrt{|x_i|})$ & 30 & $[-500, 500]$ \\
$F_9$ & $\sum_{i=1}^{n}[x_i^2 - 10\cos(2\pi x_i) + 10]$ & 30 & $[-5.12, 5.12]$ \\
$F_{10}$ & $-20\exp(-0.2\sqrt{\frac{1}{n}\sum x_i^2}) - \exp(\frac{1}{n}\sum \cos(2\pi x_i)) + 20 + e$ & 30 & $[-32, 32]$ \\
$F_{11}$ & $\frac{1}{4000}\sum x_i^2 - \prod \cos(\frac{x_i}{\sqrt{i}}) + 1$ & 30 & $[-600, 600]$ \\
$F_{12}$ & $\frac{\pi}{n}\{10\sin^2(\pi y_1) + \sum (y_i-1)^2 [1+10\sin^2(\pi y_{i+1})] + (y_n-1)^2\} + \sum u(x_i,10,100,4)$ & 30 & $[-50, 50]$ \\
$F_{13}$ & $0.1\{\sin^2(3\pi x_1) + \sum (x_i-1)^2 [1+\sin^2(3\pi x_{i+1})] + (x_n-1)^2 [1+\sin^2(2\pi x_n)]\} + \sum u(x_i,5,100,4)$ & 30 & $[-50, 50]$ \\
$F_{14}$ & $(\frac{1}{500} + \sum_{j=1}^{25} \frac{1}{j + \sum_{i=1}^2 (x_i - a_{ij})^6})^{-1}$ & 2 & $[-65, 65]$ \\
$F_{15}$ & $\sum_{i=1}^{11} [a_i - \frac{x_1(b_i^2 + b_ix_2)}{b_i^2 + b_ix_3 + x_4}]^2$ & 4 & $[-5, 5]$ \\
\bottomrule
\end{tabularx}
\end{table}

\begin{table}[htbp]
\centering
\caption{Ablation results of INGO and its variants on benchmark functions.}
\label{tab:ablation_results}
\footnotesize 
\begin{tabularx}{\columnwidth}{@{} l l X X X X @{}}
\toprule
\textbf{Func.} & \textbf{Metric} & \textbf{NGO} & \textbf{INGO-DCMIS} & \textbf{INGO-BPED} & \textbf{INGO} \\ 
\midrule
\multirow{2}{*}{$F_1$} & Mean & 1.27e-87 & 2.99e-87 & 1.94e-82 & 4.22e-82 \\
                    & Std  & 1.87e-87 & 7.76e-87 & 4.29e-82 & 1.51e-81 \\ \addlinespace
\multirow{2}{*}{$F_2$} & Mean & 1.35e-45 & 1.17e-45 & 4.64e-43 & 1.30e-43 \\
                    & Std  & 1.89e-45 & 1.11e-45 & 1.27e-42 & 2.11e-43 \\ \addlinespace
\multirow{2}{*}{$F_3$} & Mean & 1.48e-22 & 3.02e-23 & 4.52e-47 & \textbf{4.93e-47} \\
                    & Std  & 5.66e-22 & 6.79e-23 & 2.09e-46 & 1.89e-46 \\ \addlinespace
\multirow{2}{*}{$F_4$} & Mean & 2.06e-37 & 2.34e-37 & 1.25e-35 & 1.16e-35 \\
                    & Std  & 2.09e-37 & 2.49e-37 & 2.15e-35 & 1.58e-35 \\ \addlinespace
\multirow{2}{*}{$F_5$} & Mean & 25.9655  & 25.9506  & 25.5573  & \textbf{25.5184} \\
                    & Std  & 0.4854   & 0.4327   & 0.8491   & 0.7869 \\ \addlinespace
\multirow{2}{*}{$F_6$} & Mean & 6.41e-04 & 0.0012   & 9.17e-07 & 1.95e-06 \\
                    & Std  & 0.0017   & 0.0044   & 3.81e-07 & 5.13e-06 \\ \addlinespace
\multirow{2}{*}{$F_7$} & Mean & 6.18e-04 & 7.16e-04 & 3.21e-04 & \textbf{2.83e-04} \\
                    & Std  & 2.48e-04 & 2.66e-04 & 1.45e-04 & 1.69e-04 \\ \addlinespace
\multirow{2}{*}{$F_8$} & Mean & -7.48e+03& -7.61e+03& -7.53e+03& -7.48e+03 \\
                    & Std  & 385.1854 & 465.8406 & 707.7623 & 835.0605 \\ \addlinespace
\multirow{2}{*}{$F_9$} & Mean & \textbf{0} & \textbf{0} & 3.74e-07 & 7.60e-11 \\
                    & Std  & 0        & 0        & 2.05e-06 & 2.43e-10 \\ \addlinespace
\multirow{2}{*}{$F_{10}$} & Mean & 6.01e-15 & 6.01e-15 & 1.63e-15 & 1.98e-15 \\
                    & Std  & 1.79e-15 & 1.79e-15 & 1.70e-15 & 1.79e-15 \\ \addlinespace
\multirow{2}{*}{$F_{11}$} & Mean & \textbf{0} & \textbf{0} & \textbf{0} & \textbf{0} \\
                    & Std  & 0        & 0        & 0        & 0 \\ \addlinespace
\multirow{2}{*}{$F_{12}$} & Mean & 6.91e-06 & 5.93e-05 & 7.47e-08 & \textbf{6.81e-08} \\
                    & Std  & 1.24e-05 & 2.40e-04 & 2.96e-08 & 3.23e-08 \\ \addlinespace
\multirow{2}{*}{$F_{13}$} & Mean & 0.2542   & 0.2355   & 0.0187   & 0.0195 \\
                    & Std  & 0.1984   & 0.2148   & 0.0294   & 0.0264 \\ \addlinespace
\multirow{2}{*}{$F_{14}$} & Mean & 0.9980   & 0.9980   & 0.9980   & 0.9980 \\
                    & Std  & 0        & 4.12e-17 & 0        & 5.83e-17 \\ \addlinespace
\multirow{2}{*}{$F_{15}$} & Mean & 3.08e-04 & 3.08e-04 & 4.60e-04 & \textbf{3.07e-04} \\
                    & Std  & 3.39e-07 & 2.96e-07 & 3.47e-04 & 1.47e-19 \\
\bottomrule
\end{tabularx}
\end{table}
\subsubsection{Analysis of Ablation Study Results}

The experimental results presented in Table 3 demonstrate that both proposed enhancement strategies—DCMIS and BPED—significantly bolster the optimization performance of the algorithm while exhibiting a high degree of complementarity.

The DCMIS initialization strategy effectively strengthens the global exploration capability of the algorithm. This is particularly evident in functions such as $F_2$ and $F_8$, where the best solutions identified by INGO-DCMIS are markedly superior to those of the original NGO algorithm. However, when employed in isolation, this strategy yields relatively higher standard deviations for functions such as $F_6$ and $F_{12}$, suggesting that while it enhances search breadth, it may slightly affect the stability of the optimization process.

In contrast, the BPED strategy plays a pivotal role in refining the local exploitation capacity and improving the robustness of the algorithm. For several complex multimodal functions, including $F_3$, $F_4$, $F_6$, $F_{12}$, and $F_{13}$, the integration of BPED not only elevates the convergence accuracy by several orders of magnitude but also substantially reduces the standard deviation of the optimization results. This proves that the elite-guidance and stagnant-agent reinitialization mechanisms successfully assist the algorithm in circumventing local optima while maintaining population diversity.

By synthesizing both strategies, the INGO algorithm achieves the most optimal comprehensive performance across the vast majority of test functions. Notably, for functions such as $F_9$, $F_{12}$, and $F_{15}$, INGO maintains exceptional convergence precision coupled with the highest stability. This synergy demonstrates that the collaborative effect of DCMIS and BPED effectively balances the trade-off between exploration and exploitation.

\subsubsection{Statistical Analysis via Boxplot Distributions}

\begin{figure*}[ht]
  \centering
  \includegraphics[width=\linewidth]{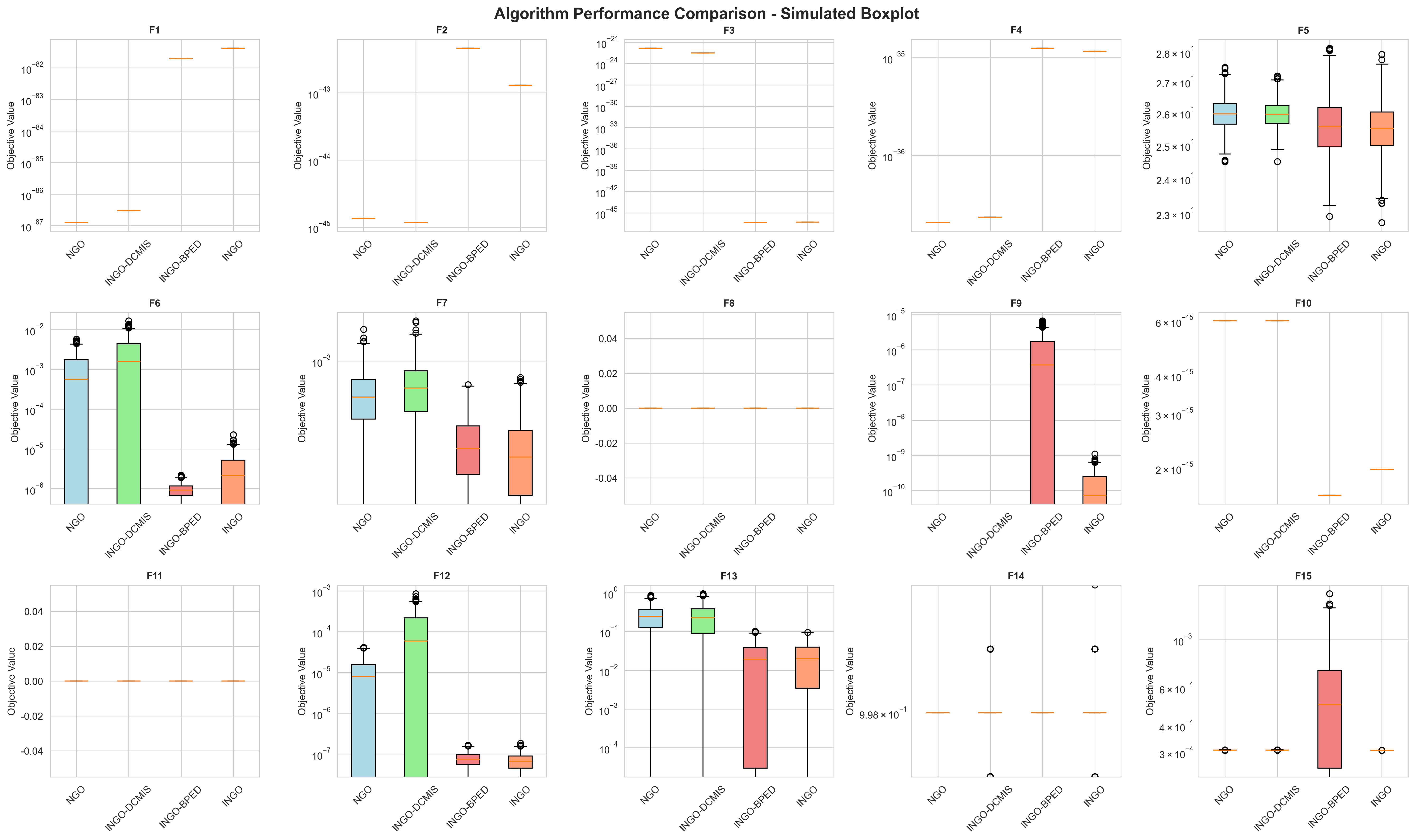}
  \caption{Algorithm Performance Comparison - Simulated Boxplot}
  \label{fig4}
\end{figure*}

Fig.~\ref{fig4} illustrates the boxplot distributions of the fitness values obtained by five competing algorithms over multiple independent trials. A comparative analysis of these distributions yields several critical insights into the optimization performance and robustness of each method:

The boxplot for the standard NGO algorithm exhibits the greatest vertical extension (length) and the highest overall position among all groups. This indicates relatively low optimization precision and poor robustness, as the fitness values are characterized by high variance and suboptimal convergence.

For the second group (INGO-DCMIS), which incorporates only the DCMIS initialization strategy, the interquartile range (IQR) is markedly narrower compared to the baseline NGO. This contraction demonstrates that the DCMIS strategy significantly enhances the stability of the algorithm by reducing the performance fluctuations typically associated with stochastic pseudo-random initialization.

In the third group (INGO-BPED), where only the BPED strategy is introduced, the median line of the boxplot is observed to shift significantly downward. This downward displacement signifies a substantial improvement in search precision, confirming that the BPED strategy effectively facilitates the escape from local optima and alleviates stagnation at local extrema during the late-stage convergence.

The final group, representing the full INGO algorithm with both DCMIS and BPED strategies, displays the most compact boxplot located at the lowest fitness level. The bottom of the distribution appears as a tight, horizontal line segment, indicating that the simultaneous integration of both strategies enables the algorithm to maintain exceptional convergence accuracy while ensuring superior stability and high robustness across all trials.

\subsection{WSN Coverage Optimization Performance and Analysis}

To comprehensively evaluate the performance of the proposed INGO algorithm, a comparative study was conducted against four prominent optimization metaheuristics: the standard NGO algorithm\cite{b22}, Artificial Bee Colony (ABC)\cite{b23}, Improved Wild Horse Optimizer (IWHO), and the Firefly Algorithm (FA)\cite{b24}.

To ensure statistical reliability and mitigate the impact of stochastic fluctuations, each algorithm was executed for 30 independent trials. The average performance metrics across these trials were utilized to assess the efficacy of each method in the context of WSN coverage optimization. The network coverage rate was quantified using the objective function $Cov$ as defined in Equation (4). All simulation parameters, including the deployment area dimensions and sensor configurations, were standardized according to the settings summarized in Table 4.

\begin{table}[htbp]
\centering
\caption{Simulation parameters for WSN coverage optimization.}
\label{tab:wsn_parameters}
\begin{tabularx}{\columnwidth}{@{} X l @{}}
\toprule
\textbf{Parameter Name} & \textbf{Value} \\ 
\midrule
Monitoring area dimensions ($L \times M$) & $50\text{ m} \times 50\text{ m}$ \\
Total number of sensor nodes ($N$)        & 35 \\
Sensing radius ($R$)                      & 5 m \\
Communication radius ($R_c$)              & 10 m \\
Discretization granularity ($\Delta$)      & 0.8 \\
Maximum iterations ($T_{max}$)            & 500 \\
Population size                     & 30 \\
Independent experimental runs             & 30 \\
\bottomrule
\end{tabularx}
\end{table}

\subsubsection{Discussion of WSN Deployment and Coverage Statistics}

\begin{figure}[ht]
  \centering
  \includegraphics[width=0.9\linewidth]{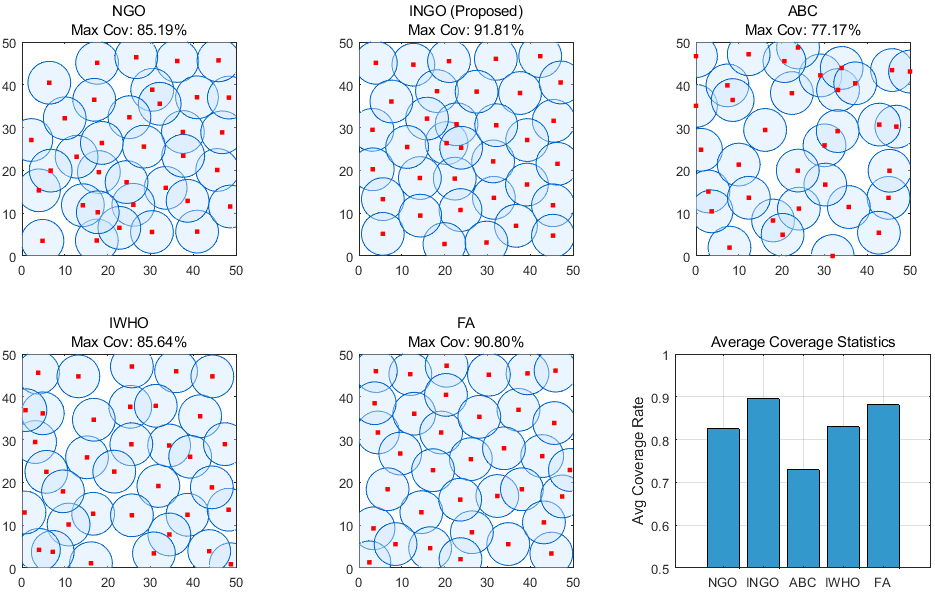}
  \caption{Wireless Sensor Network Coverage}
  \label{fig5}
\end{figure}

The node distribution maps generated by the five competing algorithms are presented in Fig.~\ref{fig5}. A comparative analysis of these spatial layouts reveals the following :

The INGO algorithm produces a significantly more uniform node distribution. Unlike the stochastic initialization of the standard NGO, the integration of the DCMIS strategy ensures that sensor nodes are evenly dispersed across the monitoring area from the onset. This effectively eliminates initial deployment imbalances.
In contrast, the ABC algorithm exhibits substantial overlapping of sensing circles (high redundancy) and leaves multiple uncovered "blind spots," particularly along the peripheral boundaries. While the NGO and IWHO algorithms achieve broad coverage, their distribution patterns lack the precision and optimized spacing demonstrated by INGO.

The statistical performance, as illustrated in the coverage histograms, further quantifies the superiority of the proposed method. The INGO algorithm achieves a peak coverage rate of 92.81\%, significantly outperforming the competing metaheuristics. The baseline NGO reaches only 85.19\%, indicating a deficiency in escaping local optima during the late-stage search. The integration of the BPED strategy empowers INGO to circumvent such stagnation, facilitating the discovery of deeper optimal solutions and extricating the population from local extrema. Although the FA algorithm achieves a respectable coverage of 91.31\%, it remains 1.5\% below the performance of INGO. Notably, INGO provides a substantial improvement of 17.63\% over the ABC algorithm's 75.18\%, further validating the potent competitiveness and robustness of the proposed enhancements in complex WSN optimization tasks.

\subsubsection{Convergence Analysis and Computational Efficiency}

\begin{figure}[ht]
  \centering
  \includegraphics[width=0.8\linewidth]{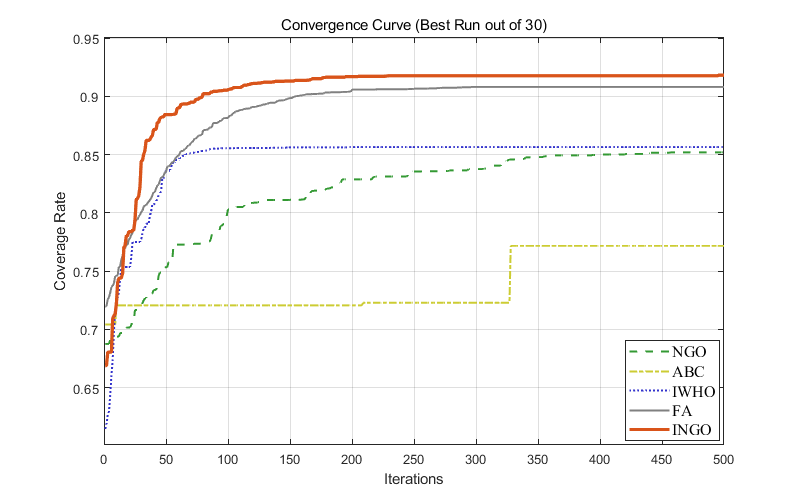}
  \caption{Convergence Curve}
  \label{fig6}
\end{figure}

The convergence characteristics of the five competing algorithms are illustrated in Fig.~\ref{fig6}. The convergence curve of INGO exhibits a rapid ascent during the initial iterations. This significantly underscores the efficacy of the DCMIS strategy, which enhances the stochasticity and uniformity of the initial population. By providing a high-quality initial distribution, DCMIS allows the algorithm to quickly locate promising regions of the search space, effectively mitigating the search blindness associated with the standard pseudo-random initialization observed in NGO and other baseline methods.

Within the first 200 iterations, the curves for NGO, ABC, and IWHO progressively plateau, indicating that these conventional algorithms have stagnated within local optima. In contrast, the INGO curve maintains a steady upward trajectory. This sustained progress is attributed to the BPED strategy, which empowers the algorithm to break through local optimal traps and continue exploring for higher fitness values.

By approximately 300 iterations, the INGO curve begins to level off, eventually stabilizing at a peak coverage rate of 92.81\%. Compared to other heuristics, ABC suffers from severe premature convergence, while FA, despite achieving relatively high precision, exhibits a significantly slower initial search speed than INGO. IWHO also fails to match the final coverage performance of the proposed method.

The results demonstrate that the INGO algorithm not only achieves superior coverage performance but also possesses exceptional search efficiency and robust global optimization capabilities.

The comparative results of the WSN coverage rates, as summarized in Table 5, further validate the substantial advantages of the proposed algorithm. The INGO algorithm achieves a remarkable average coverage rate of 91.90\%, demonstrating a clear leading edge over competing methods. Specifically, it provides significant improvements compared to the mean values of FA (88.30\%), IWHO (82.66\%), and the baseline NGO (82.51\%). These data suggest that the integration of the DCMIS and BPED strategies effectively bolsters the algorithm’s capability to escape local traps and circumvents premature convergence.

Regarding the standard deviation metric, the baseline NGO algorithm exhibits the lowest variance. However, while its fluctuations are minimal, its low average coverage (82.51\%) indicates a "low-level stagnation" phenomenon. This suggests that the original NGO is prone to getting trapped in local extrema during early iterations and lacks the exploratory drive to refine its solution.
Although the standard deviation of INGO is marginally higher than that of NGO (a negligible difference of only 0.0014), its performance fluctuates within a high-coverage interval of 87\% to 93\%. This represents a form of beneficial exploration, where the algorithm maintains moderate population vitality to rectify minor errors even as it approaches the theoretical global optimum.

Consequently, the INGO algorithm sacrifices an infinitesimal amount of stability to achieve a fundamental leap in coverage performance. It successfully realizes a high-level dynamic equilibrium, ensuring both exceptional accuracy and reliable optimization results for WSN deployment.

\begin{table}[htbp]
\centering
\caption{Statistical comparison of WSN coverage performance across different algorithms.}
\label{tab:coverage_statistics}
\begin{tabularx}{\columnwidth}{@{} l X X X X @{}}
\toprule
\textbf{Algorithm} & \textbf{Best} & \textbf{Worst} & \textbf{Average} & \textbf{Std Dev} \\ 
\midrule
NGO  & 85.19\% & 80.88\% & 82.51\% & 0.01088 \\ \addlinespace
\textbf{INGO} & \textbf{92.81\%} & \textbf{87.10\%} & \textbf{91.90\%} & 0.01223 \\ \addlinespace
ABC  & 75.18\% & 71.68\% & 72.54\% & 0.01261 \\ \addlinespace
IWHO & 84.76\% & 80.68\% & 82.66\% & 0.01420 \\ \addlinespace
FA   & 91.31\% & 85.94\% & 88.30\% & 0.01231 \\ 
\bottomrule
\end{tabularx}
\end{table}

\subsection{Connectivity Reliability Analysis}

To evaluate the communication reliability of the WSN model, the connectivity rate ($\eta_{conn}$) is employed as the primary metric. The network is modeled as an undirected graph $G=(S,E)$, where $S$ is the set of sensor nodes and E is the set of edges. An edge exists between two nodes if their Euclidean distance is less than the communication radius $R_c$. Let $C=\left\{ C_1, C_2, \dots, C_m \right\}$ represent the set of connected components in graph $G$, such that each $C_k$ is a mutually connected subset of nodes and $\bigcup_{k=1}^{m}C_k=S$. The connectivity rate $\eta_{conn}$ is defined as the ratio of the number of nodes in the largest connected component to the total number of nodes $N$, formulated as follows:

\begin{equation}
\eta_{conn} = \frac{\max_{k} |C_k|}{N} \times 100\%\label{eq}
\end{equation}

where $|C_k|$ denotes the number of nodes in the $k-th$ connected component, and $max_{k} |C_k|$ represents the size of the largest connected subgraph in the network.

\begin{figure}[ht]
  \centering
  \includegraphics[width=\linewidth]{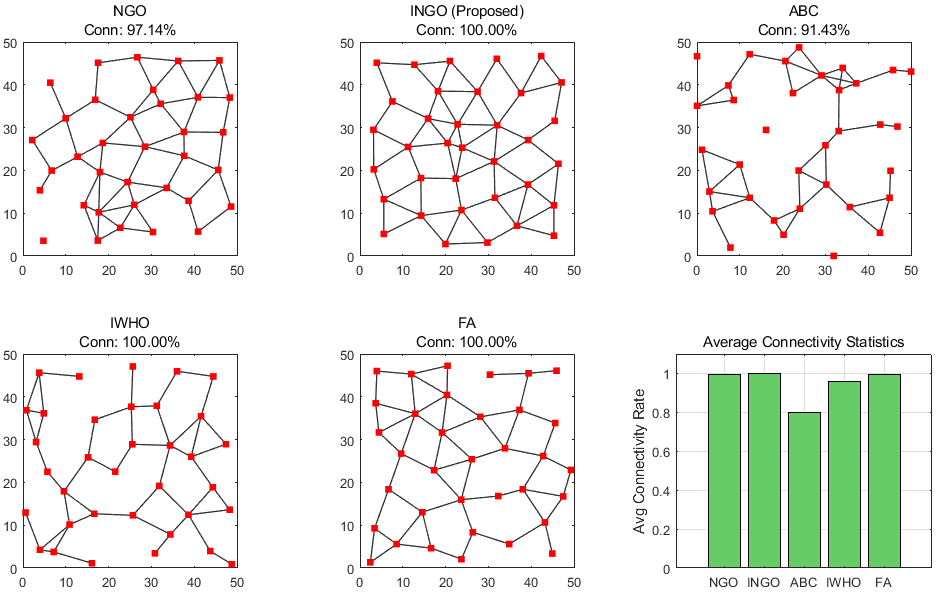}
  \caption{Algorithm Connectivity Network Structure}
  \label{fig7}
\end{figure}

The communication topologies generated by the competing algorithms are illustrated in Fig.~\ref{fig7}. From the perspective of network topology, the INGO algorithm constructs the most comprehensive and uniform communication structure. The results indicate that the NGO, IWHO, and FA algorithms all suffer from isolated nodes, leading to localized communication link fractures. In contrast, the INGO algorithm achieves 100.00\% full connectivity, forming a well-distributed and regular mesh topology. This demonstrates that while maximizing the coverage area, the proposed algorithm effectively prevents node isolation, thereby ensuring the reliability and robustness of the entire network communication.

\subsubsection{Statistical Analysis of Network Connectivity}

The statistical histograms of the connectivity rates across multiple independent runs are evaluated to further assess the stability of the generated network topologies.The column corresponding to the INGO algorithm reaches the maximum possible value and the average connectivity rate of INGO consistently maintains a peak level of 100.00\%. Such results demonstrate that the algorithm's stability is markedly superior to that of the other competing metaheuristics.

The ABC algorithm, while occasionally achieving full connectivity, exhibits the lowest average connectivity rate (approximately 75\%). This significant fluctuation suggests that the networks optimized by ABC are highly susceptible to fragmentation and topological disconnections.

Collectively,  the INGO algorithm successfully achieves an optimal equilibrium between coverage maximization and communication link maintenance. INGO proves to be a robust solution for practical WSN deployments where both sensing range and data transmission reliability are critical.

\section{Conclusion}

This paper addresses the critical challenge of coverage optimization in WSNs by proposing an Improved Northern Goshawk Optimization (INGO) algorithm incorporating multi-strategy fusion. By integrating a Diverse Chaotic Map Initialization Strategy (DCMIS), the algorithm generates an initial population characterized by superior quality and a more uniform spatial distribution compared to stochastic methods, and the integration of Bidirectional Population Evolutionary Dynamics (BPED) enables the algorithm to effectively circumvent premature convergence and escape local optima. This strategy successfully balances global exploration with local exploitation, thereby substantially improving convergence efficiency.

Comprehensive simulation experiments demonstrate that the proposed INGO algorithm consistently outperforms several prominent metaheuristics (NGO, ABC, IWHO, and FA) across key performance metrics, specifically achieving a peak coverage rate of 92.81\% and maintaining 100.00\% network connectivity. These results validate the effectiveness and superiority of the proposed enhancements in optimizing WSN node deployment.

Future research will focus on extending the application of the INGO algorithm to more complex real-world scenarios. This includes addressing three-dimensional space deployment, dynamic mobile node scheduling, and energy-efficiency balancing to further verify the universality and robustness of the algorithm in diverse and constrained environments.

\section*{Acknowledgment}
\bibliographystyle{IEEEtran}
\bibliography{Conference-LaTeX-template_10-17-19/references}

\begin{thebibliography}{10}
\providecommand{\url}[1]{#1}
\csname url@samestyle\endcsname
\providecommand{\newblock}{\relax}
\providecommand{\bibinfo}[2]{#2}
\providecommand{\BIBentrySTDinterwordspacing}{\spaceskip=0pt\relax}
\providecommand{\BIBentryALTinterwordstretchfactor}{4}
\providecommand{\BIBentryALTinterwordspacing}{\spaceskip=\fontdimen2\font plus
\BIBentryALTinterwordstretchfactor\fontdimen3\font minus
  \fontdimen4\font\relax}
\providecommand{\BIBforeignlanguage}[2]{{%
\expandafter\ifx\csname l@#1\endcsname\relax
\typeout{** WARNING: IEEEtran.bst: No hyphenation pattern has been}%
\typeout{** loaded for the language `#1'. Using the pattern for}%
\typeout{** the default language instead.}%
\else
\language=\csname l@#1\endcsname
\fi
#2}}
\providecommand{\BIBdecl}{\relax}
\BIBdecl

\bibitem{b1}
S.~P. Racharla and K.~Jeyaraj, ``A robust approach for energy-aware node
  localization in wireless sensor network using fitness-based hybrid heuristic
  algorithms,'' \emph{International Journal of Communication Systems}, vol.~38,
  no.~2, p. e6079, 2025.

\bibitem{b2}
J.~Amutha, S.~Sharma, and J.~Nagar, ``Wsn strategies based on sensors,
  deployment, sensing models, coverage and energy efficiency: Review,
  approaches and open issues,'' \emph{Wireless Personal Communications}, vol.
  111, no.~2, pp. 1089--1115, 2020.

\bibitem{b3}
K.~Gulati, R.~S.~K. Boddu, D.~Kapila, S.~L. Bangare, N.~Chandnani, and
  G.~Saravanan, ``A review paper on wireless sensor network techniques in
  internet of things (iot),'' \emph{Materials Today: Proceedings}, vol.~51, pp.
  161--165, 2022.

\bibitem{b4}
Y.~Y. Ghadi, T.~Mazhar, T.~Al~Shloul, T.~Shahzad, U.~A. Salaria, A.~Ahmed, and
  H.~Hamam, ``Machine learning solutions for the security of wireless sensor
  networks: A review,'' \emph{Ieee Access}, vol.~12, pp. 12\,699--12\,719,
  2024.

\bibitem{b5}
R.~Jia and H.~Zhang, ``Wireless sensor network (wsn) model targeting energy
  efficient wireless sensor networks node coverage,'' \emph{IEEe Access},
  vol.~12, pp. 27\,596--27\,610, 2024.

\bibitem{b6}
D.~Samiayya, J.~Daniel, and A.~Chandrasekar, ``Fuzzy trust-based hybrid levy
  snake optimization for secure and reliable wireless sensor networks,''
  \emph{International Journal of Communication Systems}, vol.~38, no.~17, p.
  e70264, 2025.

\bibitem{b7}
J.~Tang, G.~Liu, and Q.~Pan, ``A review on representative swarm intelligence
  algorithms for solving optimization problems: Applications and trends,''
  \emph{IEEE/CAA Journal of Automatica Sinica}, vol.~8, no.~10, pp. 1627--1643,
  2021.

\bibitem{b8}
W.~Li, G.-G. Wang, and A.~H. Gandomi, ``A survey of learning-based intelligent
  optimization algorithms.'' \emph{Archives of Computational Methods in
  Engineering}, vol.~28, no.~5, 2021.

\bibitem{b9}
X.~Wang, H.~Hu, Y.~Liang, and L.~Zhou, ``On the mathematical models and
  applications of swarm intelligent optimization algorithms,'' \emph{Archives
  of Computational Methods in Engineering}, vol.~29, no.~6, pp. 3815--3842,
  2022.

\bibitem{b10}
M.~Toloueiashtian, M.~Golsorkhtabaramiri, and S.~Y.~B. Rad, ``An improved whale
  optimization algorithm solving the point coverage problem in wireless sensor
  networks,'' \emph{Telecommunication Systems}, vol.~79, no.~3, pp. 417--436,
  2022.

\bibitem{b11}
J.~Akram, H.~S. Munawar, A.~Z. Kouzani, and M.~P. Mahmud, ``Using adaptive
  sensors for optimised target coverage in wireless sensor networks,''
  \emph{Sensors}, vol.~22, no.~3, p. 1083, 2022.

\bibitem{b12}
Y.~Ou, F.~Qin, K.-Q. Zhou, P.-F. Yin, L.-P. Mo, and A.~Mohd~Zain, ``An improved
  grey wolf optimizer with multi-strategies coverage in wireless sensor
  networks,'' \emph{Symmetry}, vol.~16, no.~3, p. 286, 2024.

\bibitem{b13}
L.~Cao, Y.~Yue, Y.~Cai, and Y.~Zhang, ``A novel coverage optimization strategy
  for heterogeneous wireless sensor networks based on connectivity and
  reliability,'' \emph{IEEE Access}, vol.~9, pp. 18\,424--18\,442, 2021.

\bibitem{b14}
A.~B. Alnajjar, A.~M. Kadim, R.~A. Jaber, N.~A. Hasan, E.~Q. Ahmed, M.~S.~M.
  Altaei, and A.~L. Khalaf, ``Wireless sensor network optimization using
  genetic algorithm,'' \emph{Journal of Robotics and Control (JRC)}, vol.~3,
  no.~6, pp. 827--835, 2022.

\bibitem{b15}
R.~Yarinezhad and S.~N. Hashemi, ``A sensor deployment approach for target
  coverage problem in wireless sensor networks,'' \emph{Journal of Ambient
  Intelligence and Humanized Computing}, vol.~14, no.~5, pp. 5941--5956, 2023.

\bibitem{b16}
M.~Elhoseny, A.~Tharwat, X.~Yuan, and A.~E. Hassanien, ``Optimizing k-coverage
  of mobile wsns,'' \emph{Expert Systems with Applications}, vol.~92, pp.
  142--153, 2018.

\bibitem{b17}
T.~G. Nguyen, C.~So-In, N.~G. Nguyen, and S.~Phoemphon, ``A novel
  energy-efficient clustering protocol with area coverage awareness for
  wireless sensor networks,'' \emph{Peer-to-Peer Networking and Applications},
  vol.~10, no.~3, pp. 519--536, 2017.

\bibitem{b18}
M.~Dehghani, {\v{S}}.~Hub{\'a}lovsk{\`y}, and P.~Trojovsk{\`y}, ``Northern
  goshawk optimization: a new swarm-based algorithm for solving optimization
  problems,'' \emph{Ieee Access}, vol.~9, pp. 162\,059--162\,080, 2021.

\bibitem{b19}
Y.~Liang, X.~Hu, G.~Hu, and W.~Dou, ``An enhanced northern goshawk optimization
  algorithm and its application in practical optimization problems,''
  \emph{Mathematics}, vol.~10, no.~22, p. 4383, 2022.

\bibitem{b20}
H.~T. Sadeeq and A.~M. Abdulazeez, ``Improved northern goshawk optimization
  algorithm for global optimization,'' in \emph{2022 4th international
  conference on advanced science and engineering (ICOASE)}.\hskip 1em plus
  0.5em minus 0.4em\relax IEEE, 2022, pp. 89--94.

\bibitem{b21}
L.~Zeng, M.~Hu, C.~Zhang, Q.~Yuan, and S.~Wang, ``A multi-strategy-improved
  northern goshawk optimization algorithm for global optimization and
  engineering design.'' \emph{Computers, Materials \& Continua}, vol.~80,
  no.~1, 2024.

\bibitem{b22}
O.~I. Khalaf, G.~M. Abdulsahib, and B.~M. Sabbar, ``Optimization of wireless
  sensor network coverage using the bee algorithm.'' \emph{J. Inf. Sci. Eng.},
  vol.~36, no.~2, pp. 377--386, 2020.

\bibitem{b23}
C.~Zeng, T.~Qin, W.~Tan, C.~Lin, Z.~Zhu, J.~Yang, and S.~Yuan, ``Coverage
  optimization of heterogeneous wireless sensor network based on improved wild
  horse optimizer,'' \emph{Biomimetics}, vol.~8, no.~1, p.~70, 2023.

\bibitem{b24}
W.~Liu, P.~Li, Z.~Ye, and S.~Yang, ``A node deployment optimization method of
  wireless sensor network based on firefly algorithm,'' in \emph{2021 IEEE 4th
  International Conference on Advanced Information and Communication
  Technologies (AICT)}.\hskip 1em plus 0.5em minus 0.4em\relax IEEE, 2021, pp.
  167--170.

\bibitem{b25}
S.~S. Kagi and S.~V. Mallapur, ``Wireless sensor network coverage of improved
  sea lion algorithm,'' \emph{International Journal of Communication Systems},
  vol.~37, no.~18, p. e5953, 2024.

\bibitem{b26}
K.~Subburathinam, V.~Bakthavatchalam, R.~K.~C. Pandian, and K.~M. Subramaniam,
  ``Enhancing wireless sensor network connectivity and coverage using hybrid
  gwo-hsa algorithm,'' \emph{International Journal of Communication Systems},
  vol.~37, no.~14, p. e5858, 2024.

\bibitem{b27}
P.~Jangir, A.~E. Ezugwu, K.~Saleem, Arpita, S.~P. Agrawal, S.~B. Pandya,
  A.~Parmar, G.~Gulothungan, and L.~Abualigah, ``A hybrid mutational northern
  goshawk and elite opposition learning artificial rabbits optimizer for pemfc
  parameter estimation,'' \emph{Scientific Reports}, vol.~14, no.~1, p. 28657,
  2024.

\bibitem{b28}
X.~Wang, ``An intensified northern goshawk optimization algorithm for solving
  optimization problems,'' \emph{Engineering Research Express}, vol.~6, no.~4,
  p. 045267, 2024.

\bibitem{b29}
P.~D. Kusuma and F.~C. Hasibuan, ``Best-worst northern goshawk optimizer: a new
  stochastic optimization method.'' \emph{International Journal of Electrical
  \& Computer Engineering (2088-8708)}, vol.~13, no.~6, 2023.

\bibitem{b30}
S.~M. Basha, P.~Mathivanan, and A.~B. Ganesh, ``Bit level color image
  encryption using logistic-sine-tent-chebyshev (lstc) map,'' \emph{Optik},
  vol. 259, p. 168956, 2022.

\bibitem{b31}
Q.~Wu, ``Cascade-sine chaotification model for producing chaos,''
  \emph{Nonlinear Dynamics}, vol. 106, no.~3, pp. 2607--2620, 2021.

\end{thebibliography}
\end{document}